\documentclass[runningheads]{llncs}
\usepackage{graphicx}
\usepackage{subfigure}
\usepackage{amsfonts}
\usepackage{multirow}
\usepackage[colorlinks, linkcolor=red,anchorcolor=blue,citecolor=green]{hyperref}
\usepackage[misc,geometry]{ifsym} 

\begin{document}

\title{Dual Adaptive Pyramid Network for Cross-Stain Histopathology Image Segmentation}
\titlerunning{Dual Adaptive Pyramid Network for Histopathology Image Segmentation}
% \author{Paper ID: 1740}

\makeatletter
\newcommand{\printfnsymbol}[1]{%
  \textsuperscript{\@fnsymbol{#1}}%
}
\makeatother

\author{Xianxu Hou\inst{1,2}\thanks{Equal contribution} \and
Jingxin Liu\inst{1,2,3}\printfnsymbol{1}\textsuperscript{(\Letter)} \and
Bolei Xu\inst{1,2} \and
Bozhi Liu\inst{1,2} \and
Xin Chen\inst{4} \and
Mohammad Ilyas\inst{5} \and
Ian Ellis\inst{5} \and
Jon Garibaldi\inst{4} \and
Guoping Qiu\inst{1,2,4}
}
% index{Hou, Xianxu}
% index{Liu, Jingxin}
% index{Xu, Bolei}
% index{Liu, Bozhi}
% index{Chen, Xin}
% index{Ilyas, Mohammad}
% index{Ellis, Ian}
% index{Garibaldi, Jon}
% index{Qiu, Guoping}

% %
\authorrunning{X. Hou et al.}
\institute{College of Information Engineering, Shenzhen University, Shenzhen, China \and
Guangdong Key Laboratory of Intelligent Information Processing, Shenzhen University, Shenzhen, China \and
Histo Pathology Diagnostic Center, Shanghai, China \and
School of Computer Science, University of Nottingham, Nottingham, United Kingdom \and
School of Medicine, University of Nottingham, Nottingham, United Kingdom \\
\email{jingxin.liu@outlook.com}}

\maketitle              % typeset the header of the contribution

\begin{abstract}
% Deep convolutional neural network based recognition and segmentation rely on the identical distributed training data and its corresponding pixel-level labelled ground truth. Since collecting and labelling medical data are widely regarded as tedious and time consuming processes, the algorithms that able to recover the performance of the pre-trained network to new unlabelled dataset are highly desired. In this paper, domain adaptation with dual adversarial learning modules is utilized to adapt the deep segmentation network trained on source dataset to target dataset domain. In particular, semantic adaptation is to shift the source image domain to the target image domain; while structural adaptation is used considering the spatial similarities between two different domains. We evaluate our new approach on two pairs of datasets of MRI and pathological images. We show that the proposed method outperforms other state-of-the-art models.

Supervised semantic segmentation normally assumes the test data being in a similar data domain as the training data. However, in practice, the domain mismatch between the training and unseen data could lead to a significant performance drop. Obtaining accurate pixel-wise label for images in different domains is tedious and labor intensive, especially for histopathology images. In this paper, we propose a dual adaptive pyramid network (DAPNet) for histopathological gland segmentation adapting from one stain domain to another. We tackle the domain adaptation problem on two levels: 1) the image-level considers the differences of image color and style; 2) the feature-level addresses the spatial inconsistency between two domains. The two components are implemented as domain classifiers with adversarial training. We evaluate our new approach using two gland segmentation datasets with H\&E and DAB-H stains respectively. The extensive experiments and ablation study demonstrate the effectiveness of our approach on the domain adaptive segmentation task. We show that the proposed approach performs favorably against other state-of-the-art methods.

\end{abstract}

\section{Introduction}
Deep convolutional neural networks (DCNNs) have achieved remarkable success in the field of medical image segmentation \cite{litjens2017survey}, which aims to identify and segment specific regions, such as organs or lesions in MR images, and cellular structures or tumor regions in pathological images. Although excellent performance has been achieved on benchmark dataset, deep segmentation models have poor generalization capability to unseen datasets \cite{tzeng2017adversarial} due to the domain shift between the training and test data.

\begin{figure}[t]
	\setlength{\abovecaptionskip}{0.cm}
	\setlength{\belowcaptionskip}{-0.3cm}
    \centering  
    \includegraphics[height=2.8cm]{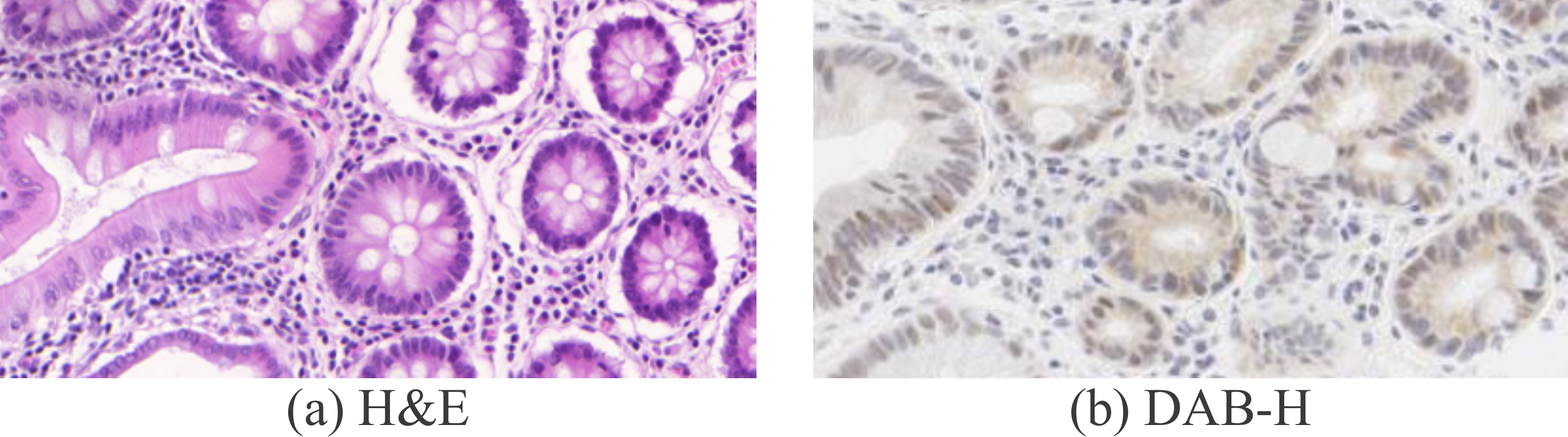} 
    \caption{Image examples of different histopathological stains. (a) Hematoxylin and Eosin; (b) Diaminobenzidene and Hematoxylin.}
    \label{fig:stainExample}
\end{figure}

Such domain shift is commonly observed especially in histopathology image analysis. For instance, the Hematoxylin and Eosin (H\&E) stained colon image has significantly different visual appearances from that stained by Diaminobenzidene and Hematoxylin (DAB-H) (Fig. \ref{fig:stainExample}). Thus, the model trained on one (source) dataset would not generalize well when applied to the other (target) dataset. Although fine-tuning the model with labelled target data could possibly alleviate the impact of domain shift, manually annotating is a time-consuming, expensive and subjective process in medical area. Therefore, it is of great interest to develop algorithms to adapt segmentation models from a source domain to a visually different target domain without requiring additional labels in the target domain.

% Such domain shift between different datasets is more common and severe especially in medical image analysis. For instance, the cardiac magnetic resonance (MR) images acquired by different scanners presents extreme different luminance histograms and noise levels as shown in Fig.1; while the Hematoxylin and Eosin (H\&E) stained colon histopathology image has significantly different visual appearance from that stained by Diaminobenzidene and Hematoxylin (DAB-H). The network trained on one (source) dataset would unsurprisingly fail when test on the other (target). The traditional approach to recover the performance is fine-tune the network with labelled target data. However, manually annotating data is a obviously time consuming, imprecise and subjective process in medical area.

Domain adaptation algorithms have been developed to address the domain-shift problem. The main insight behind these methods is trying to align visual appearance or feature distribution between the source and target domains. Zhang \textit{et al}. \cite{zhang2018fully}  render the source image with the target domain ``style'', and then learn domain-invariant representations in an adversarial manner. AdapSeg \cite{tsai2018learning} is developed to align the two domain images in the structured output space. CyCADA \cite{hoffman2018cycada} unifies adversarial  adaptation methods together with cycle-consistent image translation techniques.

% Unsupervised domain adaptation has been proposed to mapping images between domain without labelled target data. The most straightforward idea is to transfer the source domain images to target domain before segmentation. 

In this paper, we propose a DCNN-based domain adaptation algorithm for histopathology image segmentation, referred to as Dual Adaptive Pyramid Network (DAPNet). The proposed DAPNet is designed to reduce the discrepancy between two domains by incorporating two domain adaptation components on image level and feature level. The image-level adaptation considers the overall difference between source and target domain like image color and style, while feature-level adaptation addresses the spatial inconsistency of the two domains. In particular, each component is implemented as a domain classifier with an adversarial training strategy to learn domain-invariant features.

The contribution of this work can be summarized as follows. First, we develop a deep unsupervised domain adaptation algorithm for histopathology image segmentation. Second, we propose two domain adaptation components to alleviate the domain discrepancy at the image and feature levels based on pyramid features. Third, we conduct extensive experiments and our proposed DAPNet outperforms other state-of-the-art methods.

% two adversarial learning based adaptation module are designed, which are able to effectively unify the semantic context and spatial information between source and target images. Third, we conduct extensive experiments, and our proposed DAPNet outperforms other state-of-the-art models.
\begin{figure}[!tb]
	\setlength{\belowcaptionskip}{-0.2cm}
	\begin{center}
		\includegraphics[width=0.8\textwidth]{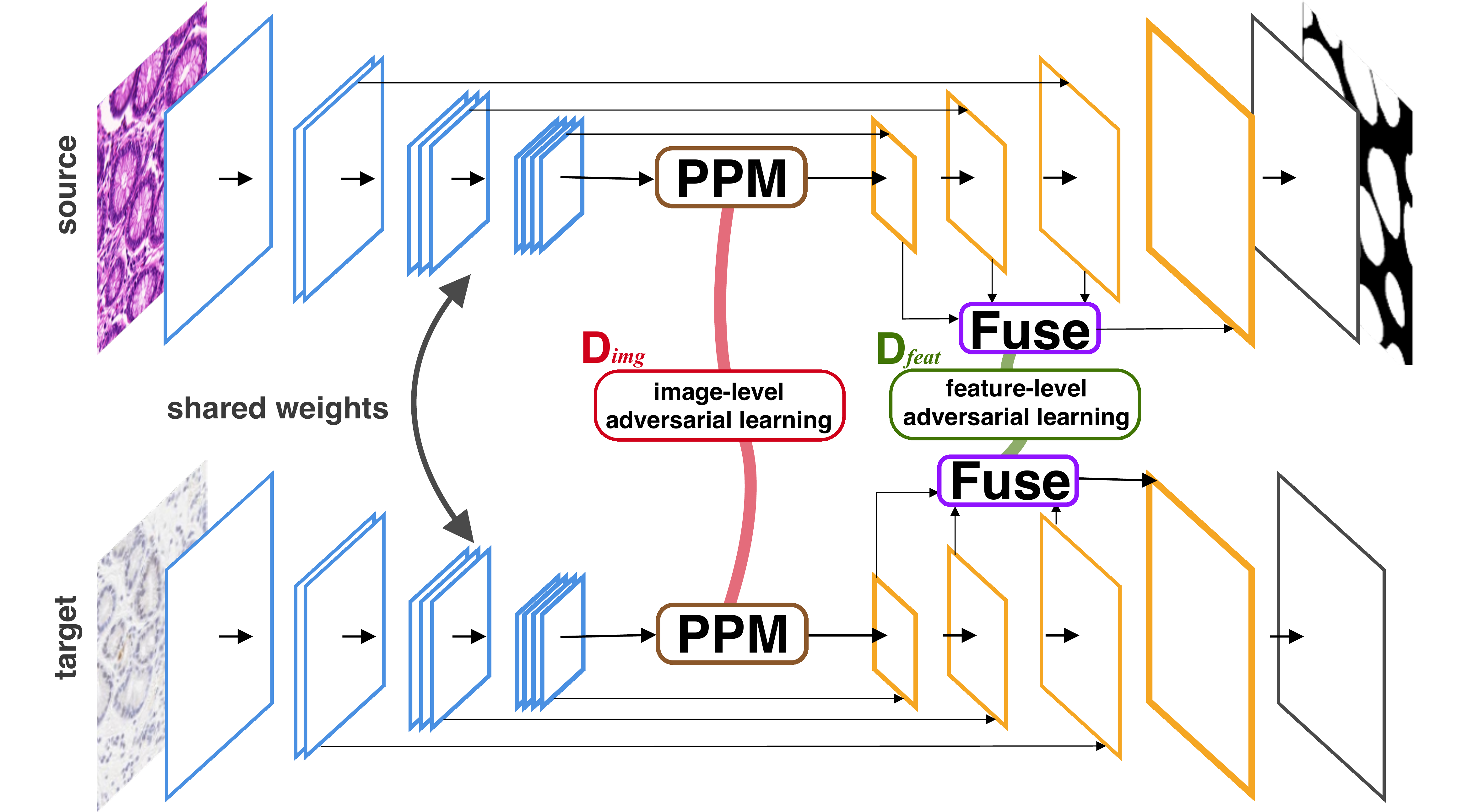}
	\end{center} 
	\caption{Overview of our DAPNet. Both source and target domain images are fed to the segmentation network. The training procedure optimizes the segmentation loss based on the source ground truth, and two domain classification losses of image-level and feature-level adversarial learning modules to make the segmentation output close to the image labels of the source domain. 
	}
	\label{fig:overview}
\end{figure}

\section{Method}
In this work, we aim to learn gland segmentation model from images with a certain stain type and apply the learned model to a different stain scenario. The training data is used as the source domain $\mathcal{S}$ while the test data with a different stain type is regarded as the target domain $\mathcal{T}$. In the $\mathcal{S}$ domain, we have access to the stained images $X_S$ as well as the corresponding ground-truth labels $Y_S$. In the target domain $\mathcal{T}$, we only have the unlabelled stained images $X_T$.

\subsection{Model Overview}
The overview of the proposed DAPNet is illustrated in Fig. \ref{fig:overview}. It contains a semantic segmentation network $G$ and two adversarial learning modules $D_{img}$ and $D_{feat}$. During training, both the source images $x_s$ and target images $x_t$ are fed into the network $G$ as inputs. The source images and the corresponding labels are used to optimize $G$ for the segmentation task, while both source and target images are used for optimizing domain adaptation losses by adversarial learning with $D_{img}$ and $D_{feat}$. %Therefore, the $G$ is encouraged to generate similar image representation and segmentation distributions for images in the target domain.

\subsection{Segmentation Network}
As shown in Fig. \ref{fig:overview}, our segmentation network consists of 3 components. First a dilated ResNet-18 \cite{he2016deep} is used as backbone to encode the input images. In order to achieve larger receptive field of our model, we apply a Pyramid Pooling Module (PPM) from PSPNet \cite{zhao2017pyramid} on the last layer of the backbone network. The PPM separates the feature map into different pooled representations with varied pyramid levels. The different levels of features are then upsampled and concatenated as the pyramid pooling global feature. Furthermore, we adopt skip connections from U-Net \cite{ronneberger2015u} and a pyramid feature fusion architecture to achieve final segmentation. The decoded feature maps are upsampled to the same spatial resolution and merged by concatenation in a pyramidal way. The output feature maps undergo a $1\times1$ convolutional layer to reduce the dimension of channel to 512. Our method involves downsampling pyramid feature extraction and upsampling pyramid feature fusion. However, the CyCADA needs to first map source training data into the target domain in pixel level.

The segmentation task is learned by minimizing both standard cross-entropy loss and Dice coefficient for images from the source domain:
\begin{equation}
\mathcal{L}_{seg} = \mathbb{E}_{x_s \sim X_S}[-y_slog(\widetilde{y}_s)] + \alpha \mathbb{E}_{x_s \sim X_S}[-\frac{2y_s\widetilde{y}_s}{y_s + \widetilde{y}_s}]
\end{equation}
where $y_s$ stands for ground-truth labels, $\widetilde{y}_s$ stands for predicted labels and $\alpha$ is the trade-off parameter.

\subsection{Domain Adaptation}

\textbf{Image-level Adaptation.}
In this work, image-level representation refers to the PPM outputs of the segmentation network $G$. Image-level adaptation helps to reduce the shift by the global image difference such as image color and image style between the source and target domains. To eliminate the domain distribution mismatch, we employ a discriminator $D_{img}$ to distinguish PPM features between source images and target images. At the same time, $D_{img}$ also guides the training of segmentation network in an adversarial manner. In particular, we employ PatchGAN \cite{isola2017image}, a fully convolutional neural operating on image patches, from which we can get a two-dimensional feature map as the discriminator outputs. The loss for training $D_{img}$ is formulated as follows:
\begin{equation}
\mathcal{L}_{img} = \mathbb{E}_{x_t \sim X_T}[log D_{img}(p_t)] +\mathbb{E}_{x_s \sim X_S}[log(1 - D_{img}(p_s))]
\end{equation}
where $p_s$ and $p_t$ denote the PPM outputs of the segmentation network $G$ for source domain and target domain.

\textbf{Feature-level Adaptation.} The feature-level representation refers to the fused feature maps before feeding into the final segmentation classifier. Aligning the feature-level representations helps to reduce the segmentation differences in both global layout and local context. Similar to image-level adaptation, we also train a domain classifier $D_{feat}$ formulated as a PatchGAN to align the feature-level distribution. Let us denote the final fused feature representation as $f_s$ and $f_t$ for source domain and target domain respectively. The loss for $D_{feat}$ is written as follows:

\begin{equation}
\mathcal{L}_{feat} = \mathbb{E}_{x_t \sim X_T}[log D_{feat}(f_t)] +\mathbb{E}_{x_s \sim X_S}[log(1 - D_{feat}(f_s))]
\end{equation}

\subsection{Overall Training Objective}
We integrate the segmentation module for source images and the two domain adaptation modules to train all the networks $G$, $D_{img}$ and $D_{feat}$ jointly. The overall objective function can be formulated as follows:
\begin{equation}
\min_{G}\max_{D_{img},D_{feat}} \mathcal{L}_{seg}(x_s, y_s) + \lambda_1\mathcal{L}_{img}(x_s, x_t) + \lambda_2\mathcal{L}_{feat}(x_s, x_t) 
\end{equation}
where $\lambda_1$ and $\lambda_2$ are two trade-off parameters. The min-max game is optimized by adversarial training and $G$ is used to achieve segmentation for images in target domain during test.

\section{Experiments and Results}
\subsection{Datasets}
\label{sec:dataset}
Two colorectal cancer gland segmentation datasets with different stains are used to evaluate our model. \textbf{Warwick-QU} dataset \cite{sirinukunwattana2017gland}, introduced in gland segmentation challenge in MICCAI 2015, consists of 165 H\&E stained images cropped from whole slide images (WSIs). The WSIs are acquired in $20 \times$ optical magnification. In our experiments, the dataset is separated into training and test sets with 85 and 80 images respectively. \textbf{GlandVision} dataset \cite{BMVC.26.42} contains 20 DAB-H stained colon images with size of $1280\times1024$, which were captured with $10 \times$ optical magnification. We randomly select 14 images for training and the rest for test. It is noted that those two datasets are labelled with different strategies. The masks in Warwick-QU cover the whole glandular structures, while GlandVision only considers the lumen regions. 

\subsection{Implementation details}
Our DAPNet employs $3 \times 3$ kernel for convolutional operations followed by a batch normalization layer. We train all the models using Adam  optimization with a batch size of 4 for 300 epochs. We randomly crop image patches of size $256 \times 256$ for training. The initial learning rate is $10^{-3}$, which is kept the same for the first 150 epochs and linearly decayed to zero over the next 150 epochs. The hyper-parameters $\alpha$, $\lambda_1$ and $\lambda_2$ are set to 1, 0.002 and 0.005 respectively. Our method is based on LSGAN \cite{mao2017least}, which replaces the negative log likelihood objective by a least square loss. This loss achieves a more stable model training and generates higher quality results.

\subsection{Results}
We evaluate the performance of our DAPNet for gland segmentation in both adaptive directions. In particular, we denote Warwick-QU (source) to GlandVision (target) as Warwick-QU $\rightarrow$ GlandVision and vice versa, and the test images in the target domain are used for evaluation. Extensive experiments including comparisons to the state-of-the-art methods and ablation study are provided.

\begin{figure}[tb]
	\setlength{\abovecaptionskip}{0.cm}
	\setlength{\belowcaptionskip}{-0.6cm}
	\begin{center}
		\includegraphics[width=0.78\textwidth]{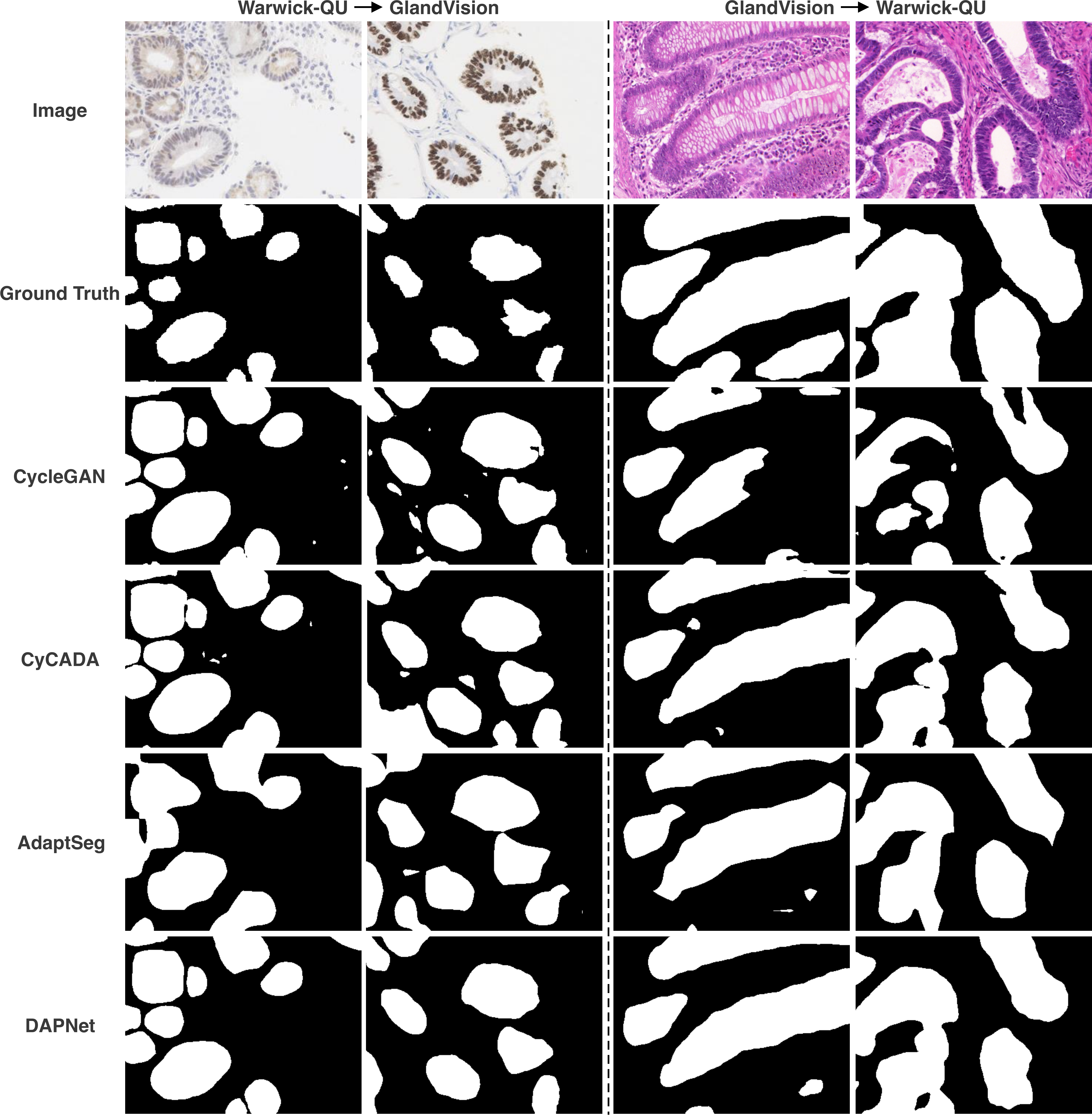}
	\end{center} 
	\caption{Qualitative results of gland segmentation adapting from Warwick-QU to GlandVision dataset (left two columns) and vice versa (right two columns).
	}
	\label{fig:compare_sota}
\end{figure}

We compare our DAPNet with three state-of-the-art unsupervised domain adaptation methods: CycleGAN \cite{zhu2017unpaired}, CyCADA \cite{hoffman2018cycada} and AdaptSeg \cite{tsai2018learning}. The comparison with CycleGAN is achieved by two stages. We first use CycleGAN transforms the source domain images to target domain, and then use the transformed images along with the corresponding label in the source domain to train the segmentation network $G$. We report the segmentation results using Pixel Accuracy (Acc.) and the Intersection over Union (IoU) in Table \ref{table:results}. We can observe that our model DAPNet outperforms all the other methods for domain adaptation between WarwickQU and GlandVision in both directions. We have repeated the model training and testing for 3 times with random parameter initializations and the same hyper-parameters. All tests have shown that our proposed method consistently outperforms other methods with statistical significance (paired t-test with p\textless 0.01). Specifically, when adapting from Warwick-QU to GlandVision, the averaged accuracy and IoU are 0.88 $\pm$ 0.0083 (Mean $\pm$ SD) and 0.68 $\pm$ 0.0021 respectively. On the other hand, the averaged accuracy and IoU are 0.76 $\pm$ 0.0105 and 0.57 $\pm$ 0.0108 respectively adapting from GlandVision to Warwick-QU. Moreover, Fig. \ref{fig:compare_sota} presents qualitative results of two example images for each of the domain adaptation case. 
%However, it can be seen in the second column of Fig. \ref{fig:compare_sota} that the adapted models can detect the mis-labelled gland structure. 
Both CycleGAN and CyCADA can successfully detect the gland structures, but the predicted masks contain irregular spot noise. AdaptSeg with only image-level adaptation can hardly segment the gland boundaries clearly. Our proposed DAPNet produces significantly better predictions with accurate layout.
\begin{table}[t]
	\setlength{\abovecaptionskip}{0.1cm}
	\setlength{\belowcaptionskip}{-0.cm}
	\centering
	\small
	\caption{Comparison with state-of-the-art methods for semantic segmentation on GlandVision adapting from Warwick-QU and vice versa.}
	\begin{tabular}{c|cc|cc} 
		\hline
		\multirow{2}{*}{method} & \multicolumn{2}{c|}{Warwick-QU $\rightarrow$ GlandVision~} & \multicolumn{2}{c}{GlandVision $\rightarrow$ Warwick-QU}  \\ 
		\cline{2-5}
		& Acc. & IoU                            & Acc. & IoU                           \\ 
		\hline
		CycleGAN  \cite{zhu2017unpaired} & 0.84      & 0.60                                & 0.74      & 0.54                               \\
		CyCADA \cite{hoffman2018cycada} & 0.84      & 0.62                                & 0.73      & 0.54                               \\
		AdapSeg \cite{tsai2018learning}  & 0.81      & 0.67                                & 0.72      & 0.52                               \\ 
		\hline
		DAPNet-NA               & 0.80      & 0.58                                & 0.73      & 0.50                               \\
		DAPNet-IA               & 0.85      & 0.60                                & 0.75      & 0.55                               \\
		DAPNet-FA               & 0.83      & 0.63                                & 0.74      & 0.53                               \\
		DAPNet                  & \textbf{0.88}    & \textbf{0.68}    & \textbf{0.76}      & \textbf{0.57}                               \\
		\hline
	\end{tabular}
	\label{table:results}
\end{table}
\begin{figure}[!t]
	\setlength{\abovecaptionskip}{-0.2cm}
	\setlength{\belowcaptionskip}{-0.4cm}
	\begin{center}
		\includegraphics[width=0.78\textwidth]{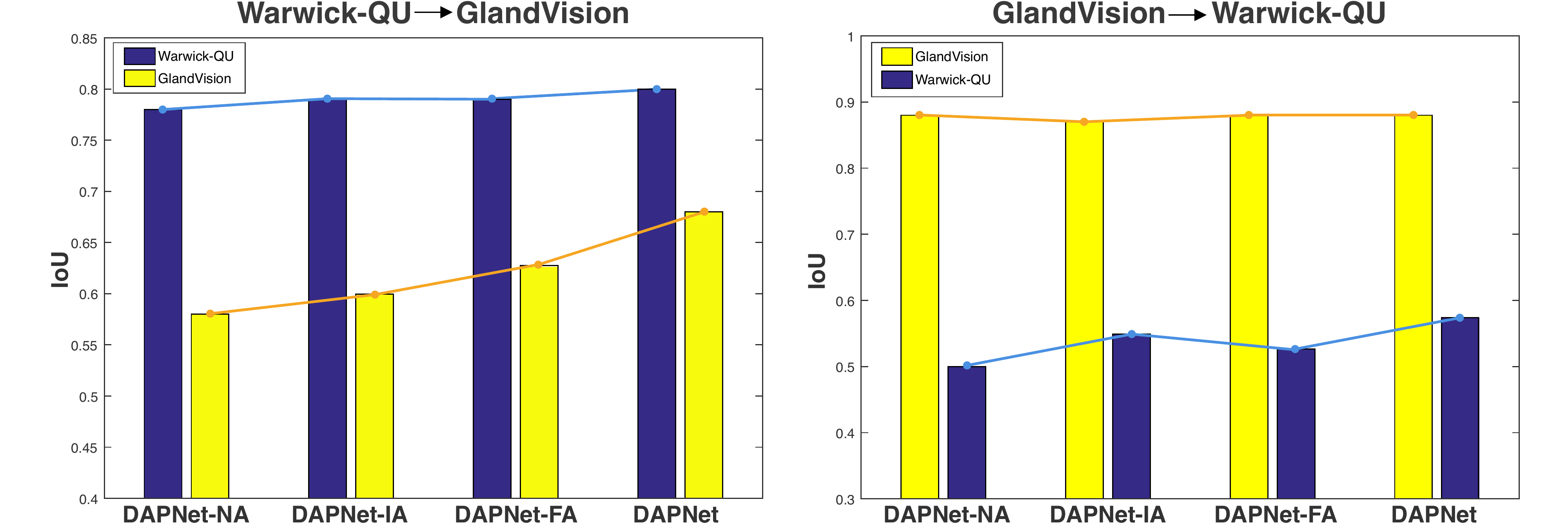}
	\end{center} 
	\caption{Performance comparison of different variants of our proposed model in terms of IoU measurements. The trained models are applied to both the source and target domain images for test. The segmentation performance for the source domain maintains at a high level while the performance of the target domain is boosted. 
	}
	\label{fig:compare_self}
\end{figure}

We further conduct ablation study to demonstrate the necessity of the two domain adaptation components of our model. In particular, we compare DAPNet with its three variants, the model trained without domain adaptation modules (DAPNet-NA), only image-level adaptation module (DAPNet-IA) and only feature-level adaptation module (DAPNet-FA). As shown in Table \ref{table:results}, we observe that the performance of the DAPNet-NA drops significantly due to the domain shift and the best results are achieved with DAPNet. It is clear that the two adaptation components can effectively alleviate the discrepancy between two domains. We also show that domain adaptation modules can boosts the segmentation performance on target domain without affecting the results on source domain (see Fig. \ref{fig:compare_self}).

\section{Conclusions}
In this paper, we study the unsupervised domain adaptive segmentation task for histopathology images. We have proposed a dual adaptive pyramid network with two domain adaptation components by adversarial training on both image and feature levels. The model is trained without target domain labels and the test procedure works as normal segmentation networks. Experimental results show that the proposed DAPNet can effectively boost the performance on unlabelled target datasets, and outperform other state-of-the-art approaches.

\small
\bibliographystyle{splncs04}
\bibliography{mybib}

\end{document}